%% file: sample-sigconf.tex
\documentclass[sigconf]{acmart}

\usepackage{booktabs} 

\setcopyright{rightsretained}

\acmDOI{10.475/123_4}

\acmISBN{123-4567-24-567/08/06}

\acmConference[KDD'18]{KDD Workshop on Machine Learning for Healthcare and Medicine}{August 2018}{London, England}
\acmYear{2018}
\copyrightyear{2018}

\begin{document}
\title[Predicting Infant Motor Development Status]{Predicting Infant Motor Development Status using Day Long Movement Data from Wearable Sensors}

\author{David Goodfellow, Ruoyu Zhi}
\orcid{1234-5678-9012}
\affiliation{%
  \institution{University of Southern California}
  \city{}
  \state{}
  \postcode{43017-6221}
}
\email{{rzhi,dgoodfel}@usc.edu}

\author{Rebecca Funke}
\affiliation{%
  \institution{University of Southern California}
  \city{}
  \state{}
  \postcode{90034}
}
\email{rfunke@usc.edu }

\author{Jos\'{e} Carlos Pulido}
\affiliation{%
  \institution{Universidad Carlos III de Madrid}
  \city{}
  \state{}
  \postcode{90034}
}
\email{jcpulido@inf.uc3m.es }

\author{Maja Matari\'{c}}
\affiliation{%
  \institution{University of Southern California}
  \city{}
  \state{}
  \postcode{90034}
}
\email{mataric@usc.edu}

\author{Beth A. Smith}
\affiliation{%
  \institution{University of Southern California}
  \city{}
  \state{}
  \postcode{90034}
}
\email{beth.smith@pt.usc.edu}

\renewcommand{\shortauthors}{D. Goodfellow et al.}

\begin{abstract}
Infants with a variety of complications at or before birth are classified as being at risk for developmental delays (AR). As they grow older, they are followed by healthcare providers in an effort to discern whether they are on a typical or impaired developmental trajectory. Often, it is difficult to make an accurate determination early in infancy as infants with typical development (TD) display high variability in their developmental trajectories both in content and timing.   Studies have shown that spontaneous movements have the potential to differentiate typical and atypical trajectories early in life using sensors and kinematic analysis systems. In this study, machine learning classification algorithms are used to take inertial movement from wearable sensors placed on an infant for a day and predict if the infant is AR or TD, thus further establishing the connection between early spontaneous movement and developmental trajectory.    
\end{abstract}

%
%



\maketitle

\input{samplebody-conf}

\bibliographystyle{ACM-Reference-Format}
\bibliography{sample-bibliography}

\end{document}

%% file: samplebody-conf.tex
\section{Introduction}
An increasing number of infants with neuromuscular impairments survive past birth due to improvements in obstetric and neonatal medicine~\cite{Alexander2003}. These infants with complications at birth are classified as being at risk for developmental delay (AR), and some are later diagnosed with cognitive and/or physical developmental delays~\cite{Ghassabian2016}. Early motor delays are often the initial signs of later developmental impairments~\cite{Ghassabian2016}. A current challenge in the field of physical therapy is identifying neuromotor impairment accurately and early enough so that interventions can take place before the developmental delay is pronounced.  One approach is tracking early spontaneous movements which have been shown to be correlated to future motor control~\cite{Piek1998}.  Specifically, researchers propose that neurological deficits could be identified by collecting high quality spontaneous movement patterns from wearable sensors and kinematic analysis systems~\cite{Groen2005, Hadders1999, Prechtl1997}.

Using sensors and kinematic analysis systems across short sessions (5-10 minutes), studies have demonstrated that kinematic variables, such as kicking frequency, spatiotemporal organization, and interjoint and interlimb coordination are different between infants with typical development (TD) and infants at risk (AR) including infants with intellectual disability ~\cite{Kouwaki2014}, myelomeningocele~\cite{Rademacher2008, Smith2011}, Down syndrome~\cite{McKay2006}, as well as infants born preterm~\cite{Geerdink1996}. 

The goal of our work is to use day-long kinematic data collected from wearable inertial sensors to predict the infant's developmental classification as TD or AR. A prediction algorithm that is able to classify between TD and AR would further establish and expand on the initial results of demonstrated differences in spontaneous movements of infants with TD and AR, supporting the potential of full-day wearable sensor data in future research and prediction algorithms. 

The remainder of the paper is organized as follows. Section 2 discusses the methodology adopted, data preprocessing, and classification. Section 3 provides the results of various methods and compares them to more interpretable methods. Section 4 discusses the future steps of this project. 

\section{Methodology}

\subsection{Data}
The dataset was provided by the Infant Neuromotor Control Laboratory in the Division of Biokinesiology and Physical Therapy at the University of Southern California. It contains day-long inertial movement data from Opal sensors affixed on each infant ankle using leg warmers with pockets. Past work has validated that these sensors can accurately record quantity of infant limb movements and do not impede or promote infant movement~\cite{smith2015daily,smithSensors}. The dataset contains 19 movement features that summarize the general movement characteristics of each infant across the time the infant was awake during that day. The dataset consists of data from 12 TD infants and 24 AR infants. Each infant was tested 3 times from 1 to 16 months (with the exception of 1 AR infant with only two sessions) at various developmental stages.  

\subsection{Data Preprocessing}
The sensors collected raw movement data from an accelerometer, gyroscope, and magnetometer. In post-analysis, we used an established algorithm that detects each leg movement from the raw sensor data~\cite{smith2017daily,smith2015daily}. Next, the movement's duration, average acceleration, peak acceleration, and type (unilateral, bilateral synchronous, or bilateral asynchronous) were determined by the algorithm. 

The dataset contains 19 movement features which can be broken down into 4 categories: movement count, duration, average acceleration, and peak acceleration. Specifically, the dataset includes the overall movement count and type of each movement for both the right and left legs. Movement count is presented as movements per hour of awake time and type is presented as a percentage of each type of movement made out of the total number of movements made. The dataset also includes the average and standard deviation of movement duration, average acceleration, and peak acceleration over all the infant's movements across the full day. Additionally, the movement features, age, developmental scale score (specifically the Alberta Infant Motor Scale \cite{AIMS}), and the AR/TD label were also included. Due to our small sample size (12 TDs and 24 ARs), we treated each new visit as a unique data record allowing for 107 samples of day-long leg movement, with 37 TDs and 100 ARs.

First, we normalized the data to correctly reflect the infants' movements due to each sensor  being active for a different amount of time during the data collection. We then split the dataset into two groups, 0 to 6 months and 6 to 12 months, because there are noticeable changes in the movement features after 6 months and important developmental milestones emerge after this time for infants~\cite{Gajewska2013}. Also, the few samples over 12 months were each from an AR infant, so we removed these samples to reduce overfitting around age. Thus, we had 16 TD and 15 AR samples for 0 to 6 months and 23 TD and 38 AR samples for 6 to 12 months. The class weights for TD and AR were balanced for each algorithm used.

\subsection{Feature Extraction}
We extracted the relevant features out of the 22 to avoid unnecessary and intercorrelated inputs. We applied three feature extraction methods (univariate feature selection~\cite{Saeys2007}, recursive feature elimination (RFE)~\cite{Mao2006}, and stepwise feature selection~\cite{McShanea1999}) separately on the 0-6 month and 6-12 month dataset. For each feature subset selected by the algorithms, we removed highly correlated features to reduce the dimensionality of the feature set.

\subsection{Classification}
We used a binary classification approach for our models. To classify infants as TD or AR, various classification algorithms were spot checked using their default parameters including Random Forest~\cite{Breiman2001}, Logistic Regression, Support Vector Machines (SVM)~\cite{Boser1992}, K-Nearest Neighbors (KNN)~\cite{Klecka1980}, Decision Trees~\cite{Quinlan1986}, and AdaBoost~\cite{Freund1999}.

We improved the performance by using grid search for tuning hyper parameters for each method, and the three top performing algorithms were put into a voting ensemble with a majority vote wins principle. The reason for ensembling is to level individual algorithm bias and to reduce overfitting by smoothing the results ~\cite{Solich1995}.

\section{Results}

\subsection{Evaluation}
We used a leave-one-out cross validation for model selection, a common approach for small datasets~\cite{Cawley2006}. Standard metrics such as accuracy, precision, recall, and F1 score were used to evaluate the classification algorithms. We focused on choosing algorithms with higher true positives (TP) and false negatives (FN) because predicting AR infants correctly is more important for initial diagnosis purposes.

\subsection{Experiments and Outcomes}
\subsubsection{0 to 6 Month Prediction Algorithms}

Based on the feature extraction methods described above, we used eight movement features: mean acceleration, peak acceleration, unilateral movements, and total movements for the right and left leg. The results from the tuned algorithms varied from 90\% average accuracy for SVM to 67.7\% for AdaBoost. To reduce overfitting, the SVM was ensembled with two other well performing algorithms: AdaBoost and Logistic Regression. The resulting ensemble received an 83.9\% accuracy with a 84\% F1 score. Every AR infant was predicted correctly. Tables 1-6 displays the outcomes of the three individual algorithms, a decision tree, the ensemble, and the baseline algorithm which is the average of ten separate weighted random predictions respectively.

\begin{table}[H]
\caption{Weighted Average Baseline}
 \vspace{-1em}
\centering
\begin{tabular}{|c|c|c|c|c|}
\hline
 Class & Accuracy & Precision & Recall & F1 Score  \\ \hline
 TD & .625 & .53 & .62 & .57 \\ \hline
 AR & .4 & .5 & .4 & .44 \\ \hline \hline 
 Average & .516 & .51 & .52 & .51 \\ \hline
 \end{tabular}
 \end{table}
 
 \vspace{-2em}
 \begin{table}[H]
 \caption{Decision Tree}
  \vspace{-1em}
\centering
 \begin{tabular}{|c|c|c|c|c|}
 \hline
  Class & Accuracy & Precision & Recall & F1 Score  \\ \hline
 TD & .75 & .75 & .75 & .75 \\ \hline
 AR & .733 & .733 & .733 & .733 \\ \hline \hline 
 Average & .742 & .742 & .742 & .742 \\ \hline
 \end{tabular}
 \end{table}
 
  \vspace{-2em}
 \begin{table}[H]
  \caption{SVM}
  \vspace{-1em}
\centering
 \begin{tabular}{|c|c|c|c|c|}
 \hline
  Class & Accuracy & Precision & Recall & F1 Score  \\ \hline
 TD & .813 & 1 & .81 & .9 \\ \hline
 AR & 1 & .83 & 1 & .91 \\ \hline \hline 
 Average & .903 & .92 & .9 & .9 \\ \hline
 \end{tabular}
 \end{table}
 
   \vspace{-2em}
 \begin{table}[H]
 \caption{AdaBoost}
  \vspace{-1em}
\centering
 \begin{tabular}{|c|c|c|c|c|}
 \hline
  Class & Accuracy & Precision & Recall & F1 Score  \\ \hline
 TD & .625 & .71 & .62 & .67 \\ \hline
 AR & .733 & .65 & .73 & .69 \\ \hline \hline 
 Average & .677 & .68 & .68 & .68 \\ \hline
 \end{tabular}
 \end{table}
 
    \vspace{-2em}
 \begin{table}[H]
 \caption{Logistic Regression}
  \vspace{-1em}
\centering
 \begin{tabular}{|c|c|c|c|c|}
 \hline
  Class & Accuracy & Precision & Recall & F1 Score  \\ \hline
 TD & .625 & .77 & .62 & .69 \\ \hline
 AR & .8 & .67 & .8 & .73 \\ \hline \hline 
 Average & .71 & .72 & .71 & .71 \\ \hline
 \end{tabular}
 \end{table}
 
     \vspace{-2em}
 \begin{table}[H]
 \caption{Ensemble:SVM, AdaBoost, and LogReg}
  \vspace{-1em}
\centering
 \begin{tabular}{|c|c|c|c|c|}
 \hline
  Class & Accuracy & Precision & Recall & F1 Score  \\ \hline
 TD & .688 & 1 & .69 & .81 \\ \hline
 AR & 1 & .5 & 1 & .86 \\ \hline \hline 
 Average & .839 & .88 & .84 & .84 \\ \hline
 \end{tabular}
 \end{table}

\subsubsection{6 to 12 Month Prediction Algorithms}

Using the features above, the same tests were conducted. The individual algorithms' accuracies ranged from 83\% for AdaBoost to 67.2\% for SVM and KNN. An ensemble was made to reduce overfitting using SVM, Adaboost, and Random Forest with majority vote as the winning prediction. The ensemble ended with a 77\% accuracy and 77\% F1 Score. Tables 7-12 list the results for the weighted average baseline, a decision tree, each individual algorithm in the ensemble, and the ensemble.

\begin{table}[H]
 \caption{Weighted Average}
  \vspace{-1em}
\centering
\begin{tabular}{|c|c|c|c|c|}
\hline
 Class & Accuracy & Precision & Recall & F1 Score  \\ \hline
 TD & .391 & .35 & .39 & .37 \\ \hline
 AR & .553 & .6 & .55 & .58 \\ \hline \hline 
 Average & .492 & .5 & .49 & .5 \\ \hline
 \end{tabular}
 \end{table}
 
 \vspace{-2em}
 \begin{table}[H]
  \caption{Decision Tree}
  \vspace{-1em}
\centering
 \begin{tabular}{|c|c|c|c|c|}
 \hline
  Class & Accuracy & Precision & Recall & F1 Score  \\ \hline
 TD & .478 & .611 & .478 & .537 \\ \hline
 AR & .816 & .721 & .816 & .765 \\ \hline \hline 
 Average & .689 & .68 & .689 & .679 \\ \hline
 \end{tabular}
 \end{table}
 
  \vspace{-2em}
 \begin{table}[H]
  \caption{SVM}
  \vspace{-1em}
\centering
 \begin{tabular}{|c|c|c|c|c|}
 \hline
  Class & Accuracy & Precision & Recall & F1 Score  \\ \hline
 TD & .609 & .56 & .61 & .58 \\ \hline
 AR & .711 & .75 & .71 & .73 \\ \hline \hline 
 Average & .672 & .68 & .67 & .67 \\ \hline
 \end{tabular}
 \end{table}
 
   \vspace{-2em}
 \begin{table}[H]
  \caption{AdaBoost}
  \vspace{-1em}
\centering
 \begin{tabular}{|c|c|c|c|c|}
 \hline
  Class & Accuracy & Precision & Recall & F1 Score  \\ \hline
 TD & .739 & .81 & .74 & .77 \\ \hline
 AR & .895 & .5 & .89 & .87 \\ \hline \hline 
 Average & .83 & .83 & .84 & .83 \\ \hline
 \end{tabular}
 \end{table}
 
    \vspace{-2em}
 \begin{table}[H]
  \caption{Random Forest}
  \vspace{-1em}
\centering
 \begin{tabular}{|c|c|c|c|c|}
 \hline
  Class & Accuracy & Precision & Recall & F1 Score  \\ \hline
 TD & .609 & .64 & .61 & .62 \\ \hline
 AR & .789 & .77 & .79 & .78 \\ \hline \hline 
 Average & .721 & .72 & .72 & .72 \\ \hline
 \end{tabular}

 \end{table}
 
      \vspace{-2em}
 \begin{table}[H]
  \caption{Ensemble:SVM,AdaBoost,Random Forest}
  \vspace{-1em}
\centering
 \begin{tabular}{|c|c|c|c|c|}
 \hline
  Class & Accuracy & Precision & Recall & F1 Score  \\ \hline
 TD & .608 & .74 & .61 & .67 \\ \hline
 AR & .868 & .79 & .87 & .82 \\ \hline \hline 
 Average & .77 & .77 & .77 & .77 \\ \hline
 \end{tabular}

 \end{table}
 
 \subsubsection{Overall Outcome}

\begin{figure}[h]
    \centering
    \includegraphics[width=.6\linewidth]{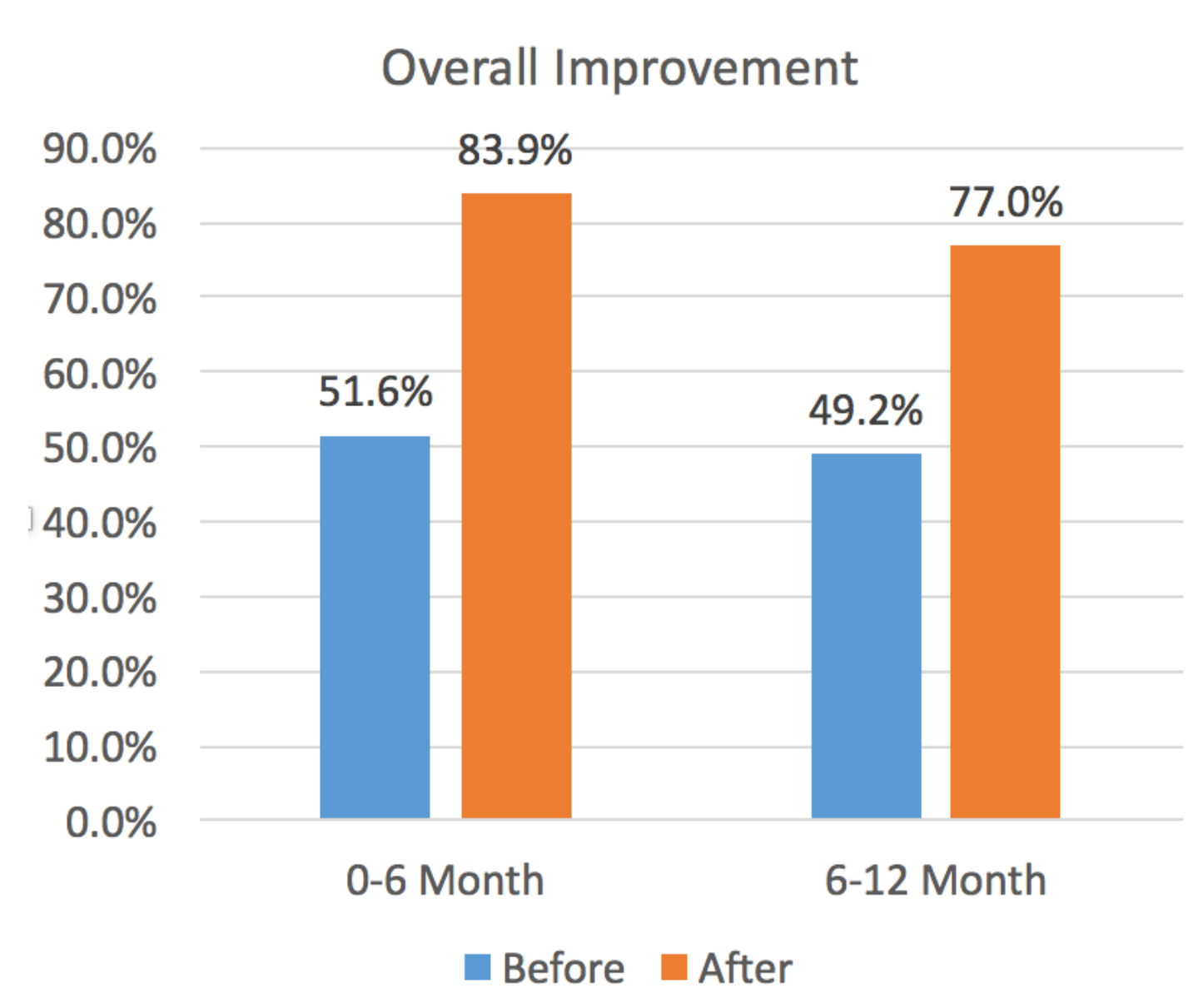}
	\caption{Comparison of accuracies between the baseline and ensemble models for 0-6 and 6-12 months.}
	\label{fig:overall}
\end{figure}

Overall, our ensemble approach in the 0-6 and 6-12 model resulted in a 32.3\% and 27.8\% increase in accuracy, respectively, when compared to the baseline prediction.  Fig. \ref{fig:overall} graphs the overall improvement in accuracy. 

\subsection{Accuracy versus Interpretability}
In selecting an ensemble model, there is an inherent trade-off between increasing accuracy and decreasing interpretability. We need our model to be as accurate as possible, but we also need interpretable results for medical professionals where interpretability is necessary for professional insight and patient and parent trust~\cite{Vellido2012}. Simpler algorithms, such as decision trees, are easily interpretable compared to ensemble models. We generated several decision trees to provide explanations to medical specialist on the importance of specific features in predicting TD and AR.

\begin{figure}[h]
    \centering
    \includegraphics[width=1\linewidth]{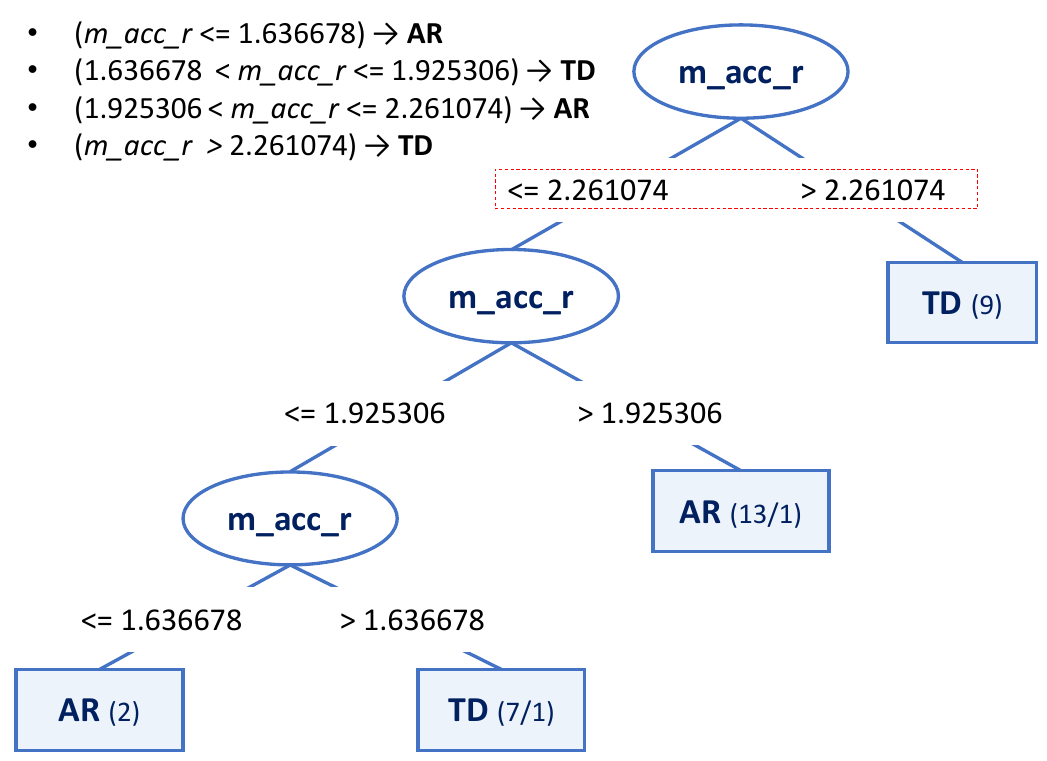}
	\caption{The top-down model generated with the 0 to 6 months dataset using J48 (C 4.5).}
	\label{fig:dt}
\end{figure}

Specifically, Fig \ref{fig:dt} shows the model decision tree resulting from executing algorithm J48 (C4.5) for the first age range (0 to 6 months). Unlike the second age range (6 to 12 months), this model obtained the  accuracy of 70.75\% with an improvement significantly better than the baseline. The peculiarity of this model is that the generated tree uses a single feature of the data set that is the mean acceleration of the right leg (m\_acc\_r). Specifically, a mean acceleration value higher than 2.261074 was always TD implying some connection with higher accelerations and typical development.

\section{Discussions}
In this work, we investigated the use of binary classification algorithms applied to day-long sensor movement data to predict typical or at risk developmental status in infants. Our analysis was split for 0-6 months and 6-12 months, and both final algorithms performed significantly better than our baseline predictions. It is clear from our analysis that from 0 to 6 months the kinematic data of AR infants are more distinct than TD, but from 6 to 12 months it becomes hard to seperate the the two classes. This could be due to the small sample size. Overall, these results further establish the relationship between the kinematic features used in the classification algorithms and infant development. 

Working with interpretable and understandable models is very important in order to obtain knowledge that can be shared with the medical experts. The top-down model generated in this work shows the importance of the mean acceleration in the prediction of AR infants. In fact, the model has found a possible threshold value (2.261074) above which typical development can be considered for an infant between 0 and 6 months. Future studies are necessary to determine the robustness of this classifier and test for relationships to developmental outcomes.

The nature of this dataset, similar to all small datasets, leads to the possibility of overfitting as well as lacking representation of the general population. When using a leave one out cross validation, the trained models may be picking up noise beyond the true signal of the pattern which could cause the model to perform poorly on new unseen instances. We paired our final ensemble model with the decision tree to provide more interpretable results for healthcare professionals.

There are several next steps we are pursuing. We would like to recruit more TD and AR infants for this study. This requires significant longitudinal interaction with each infant such as traditional developmental assessment tests, movement sensor tests, and a check-up at two years old to determine if the infant has a diagnosed developmental delay. 

Currently, each infant visit is treated independently despite having three samples per infant. A larger dataset would allow us to divide the samples by infant rather than by visit, providing more data about each infant and potentially improving our predictions. We could also create an algorithm that predicts some movement feature of an infant at visit three based on the infant's first two visits. This prediction algorithm would be helpful if an infant is participating in a new intensive intervention after their second visit, and the healthcare professional wants to compare the infant's movement results after the intervention to the infant's predicted movement pattern from the algorithm that assumes no intervention. 
 
The eventual goal is to use this methodology to predict if an at risk infant will be diagnosed with a developmental delay. Currently, developmental delays are often not diagnosed until an infant is two years old, preventing many infants from receiving early targeted interventions. Current tools, such as the AIMS score, detect early signs of atypical development, but they can only detect the extreme cases. The prediction algorithm we aim to develop would confirm that developmental delays are reflected in the movement of infants in the first few months of infancy and thus allow for more infants to receiver earlier, directed interventions.


\begin{acks}

This work was funded in part by the American Physical Therapy Association Academy of Pediatric Physical Therapy Research Grant 1 and 2 Awards (PI: Smith) and in part by NSF award 1706964 (PI: Smith, Co-PI: Matari\'c). Study procedures were approved by the Institutional Review Boards of the Oregon Health \& Science University and the University of Southern California.

\end{acks}